\def\year{2022}\relax
\documentclass[letterpaper]{article} 
\usepackage{aaai22}  
\usepackage{times}  
\usepackage{helvet}  
\usepackage{courier}  
\usepackage[hyphens]{url}  
\usepackage{graphicx} 
\usepackage{natbib}  
\usepackage{caption} 
\DeclareCaptionStyle{ruled}{labelfont=normalfont,labelsep=colon,strut=off} 
\usepackage{cite}
\usepackage{amsmath,amssymb,amsfonts}
\usepackage{graphicx}
\usepackage{textcomp}
\usepackage{xcolor}
\usepackage{multirow}
\usepackage{subfigure}
\usepackage{mathrsfs}

\usepackage{algorithm}  

\usepackage{graphicx}
\usepackage{balance} 
\usepackage{color}
\usepackage{amsmath}
\usepackage{enumitem}
\usepackage{textcomp}
\usepackage{xcolor}
\usepackage{makecell}
\usepackage{bm}
\usepackage{booktabs}
\usepackage{tabu}
\usepackage{hyperref}
\usepackage{algorithm}
\usepackage{algorithmic}

\newcommand{\sysname}{\texttt{HAGEN}}

\newtheorem{myDef}{Definition}
\newcommand{\para}[1]{\noindent{\textbf{#1}}}

\usepackage{newfloat}
\usepackage{listings}
\floatstyle{ruled}
\newfloat{listing}{tb}{lst}{}
\floatname{listing}{Listing}
\title{\sysname: Homophily-Aware Graph Convolutional Recurrent Network for Crime Forecasting}

\author{Chenyu Wang\textsuperscript{\rm 1}\footnotemark[1], Zongyu Lin\textsuperscript{\rm 1}\footnotemark[1],
Xiaochen Yang\textsuperscript{\rm 1},
Jiao Sun\textsuperscript{\rm 1},
\\
Mingxuan Yue\textsuperscript{\rm 1},
Cyrus Shahabi\textsuperscript{\rm 1}\footnotemark[2]
}
\affiliations{
  \textsuperscript{\rm 1} University of Southern California
  
}
\usepackage{bibentry}

\begin{document}

\maketitle

\begin{abstract}
The crime forecasting is an important problem as it greatly contributes to urban safety. 
Typically, the goal of the problem is to predict different types of crimes for each geographical region (like a neighborhood or censor tract) in the near future. 
Since nearby regions usually have similar socioeconomic characteristics which indicate similar crime patterns, recent state-of-the-art solutions constructed a distance-based region graph and utilized Graph Neural Network (GNN) techniques for crime forecasting, because the GNN techniques could effectively exploit the latent relationships between neighboring region nodes in the graph if the edges reveal high dependency or correlation. 
However, this distance-based pre-defined graph cannot fully capture crime correlation between regions that are far from each other but share similar crime patterns. 
Hence, to make an accurate crime prediction, the main challenge is to learn a better graph that reveals the dependencies between regions in crime occurrences and meanwhile captures the temporal patterns from historical crime records. 
To address these challenges, we propose an end-to-end graph convolutional recurrent network called \sysname \space with several novel designs for crime prediction. Specifically, our framework could jointly capture the crime correlation between regions and the temporal crime dynamics by combining an adaptive region graph learning module with the Diffusion Convolution Gated Recurrent Unit (DCGRU). Based on the homophily assumption of GNN (i.e., graph convolution works better where neighboring nodes share the same label), we propose a homophily-aware constraint to regularize the optimization of the region graph so that neighboring region nodes on the learned graph share similar crime patterns, thus fitting the mechanism of diffusion convolution. It also incorporates crime embedding to model the interdependencies between regions and crime categories. Empirical experiments and comprehensive analysis on two real-world datasets showcase the effectiveness of \sysname.
\end{abstract}

\renewcommand{\thefootnote}{\fnsymbol{footnote}}

\footnotetext[1]{Both authors contributed equally to this work. The work was conducted while the author was intern at USC’s IMSC.}
\footnotetext[2]{This is the corresponding author.}

\section{Introduction}
\label{sec.intro}

  Accurate crime forecasting can better guide the police deployment and allocation of infrastructure, resulting in great benefits for urban safety.
  Previous studies have designed several spatiotemporal deep learning frameworks for crime forecasting, including MiST~\cite{huang2019mist} and DeepCrime~\cite{huang2018deepcrime}. However, they assume a grid-based partitioning of the underlying geographic region to utilize CNN-based models which ignore the geographical extents of neighborhoods. 
  Based on the intuition that nearby regions (i.e., geographical neighborhoods which collect criminal records within a period of time) have similar crime patterns, ~\citet{sun2021crimeforecaster} leveraged a distance-based region graph to explore the connection between spatial and temporal crime patterns, and utilized a graph neural network (GNN) technique to model the spatial dependencies between regions. 
  In this setting, the crime forecasting problem is equivalent to learning a mapping function to predict the future graph signals (i.e., crime occurrences) given the historical graph signals (i.e., historical crime records) over a distance-based region graph.
  
\begin{figure}[t]
    \centering
    \includegraphics[width=0.9\linewidth]{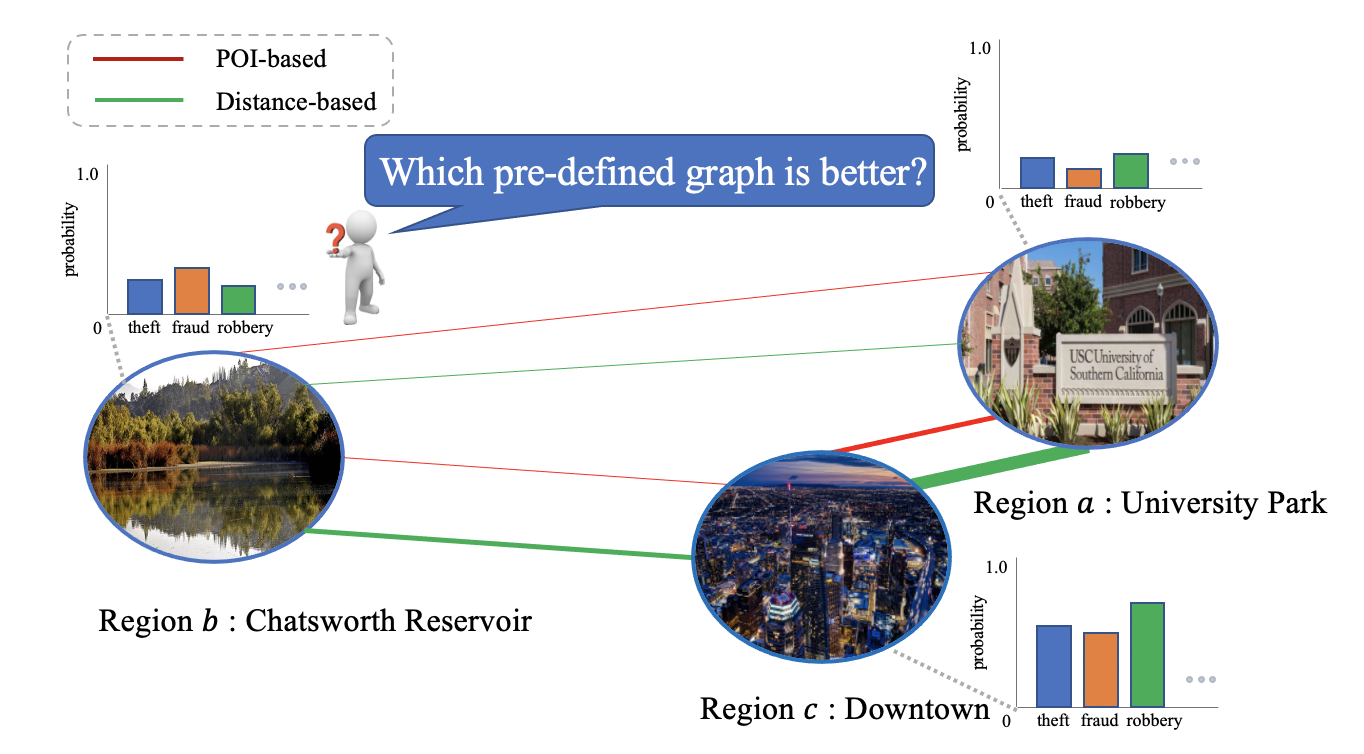}
    \caption{The width of lines represents the geographical distance or POI based similarity (the wider, the closer). As elaborated in Section \ref{sec.intro}, there exists mismatch between crime patterns and geographical distances or POI based similarity (e.g. University Park and Downtown are close to each other, while they have different crime patterns; on the other hand, although Chatworth Reservoir and University Park are far away geographically, they have similar crime patterns), which motivates our model design.
    }
    \label{fig:case}
\end{figure}

However, the geographical distance does not always reveal the real correlation of crime patterns because other shared socioeconomic factors between (possibly far apart) regions may result in similar crime patterns in the regions. 
As illustrated in Figure~\ref{fig:case}, we observe that even though the distance between regions $a$ and $b$ is much farther than that of regions $a$ and $c$, region $a$ shares similar crime patterns with $c$ but totally different from $b$. 
One may build another graph on which the edges are defined by the similarity of POIs between regions~\cite{xu2020sume}, nevertheless, the mismatching phenomenon still remains. In fact, the local government deploys heavy policing for the university in the region $a$, thus $a$ has a different crime pattern from $c$ despite both of them being urban regions.
Obviously, one can construct customized graphs for each and every case, but it is hard to predefine a graph that adapts to all cases. 

Towards this end, we propose to learn an adaptive graph that could dynamically capture the crime-specific correlations among regions from real-world data. In particular, we utilize a graph learning layer that learns the graph structure adaptively and jointly with the training process of crime forecasting.
Nevertheless, the adaptive learning of such graph structure is non-trivial and usually requires some heuristics to regularize the graph to avoid over-fitting. For example, in the traffic forecasting domain, Graph WaveNet~\cite{wu2019graph} uses the node similarity to construct the graph, and MTGNN~\cite{wu2020connecting} is the state-of-the-art spatiotemporal graph neural network that integrates a graph learning module to extract the uni-directed relations among sensors while preserving the sparsity of the graph. 
Although previous studies have considered some general graph properties in constructing the adaptive graph structure, they failed to incorporate the graph properties that are highly effective to the performance of GNN models. 
On the other hand, recent studies~\cite{zhu2020beyond,chien2021adaptive} observed that the homophily of a graph, the property that the neighboring nodes share the same labels, significantly affects the performance of GNN. 
Consider our crime forecasting task as an example, the nodes in our graph are regions and the edges between nodes should indicate a high correlation of crime patterns.
If the region nodes on the learned graph share totally different crime patterns with their neighboring nodes, GNN will aggregate and propagate the information of these dissimilar neighboring nodes resulting in the inaccurate inference of crime rates. Inspired by these studies, we would like to preserve the homophily of crime patterns among neighboring nodes of our graph.
  
In this work, we introduce \sysname, a Homophily-Aware Graph convolutional rEcurrent Network with several novel designs for crime forecasting. Specifically, we model the intra-region and inter-region dependencies of crime patterns by introducing an adaptive graph learning layer with region and crime embedding layer. 
By observing that the homophily ratio is highly correlated with the model performance,
we propose a homophily-aware constraint to boost the ability of region graph learning. Subsequently, we utilize a weighted graph diffusion layer to simulate crime diffusion and capture the crime-specific dependencies between regions. Finally, we integrate the Gated Recurrent Network in the graph diffusion layer to capture the temporal crime dynamics. Empirical results and analysis on two real-world datasets showcase the effectiveness of \sysname.
Case studies have revealed some insights for both the research community and urban planners. 
The source code and data will be published for future use by the research community. 
\section{Problem Definition}
\label{sec:problem}
In this paper, we focus on the task of crime forecasting given previous crime records. Our setting is the same as CrimeForecaster's~\cite{sun2021crimeforecaster}.

\begin{myDef}
    \textbf{Crime Record.}  
    For the inputs of forecasting model, we consider the crimes occurring in the past sequence of non-overlapping and consequent time slots $T=(t_1, \dots, t_{K})$, where $K$ is the time sequence length. For each region $r_i$, we use $\mathcal{Y}_{i}=(y_{i, 1}^{1},\dots, y_{i, l}^{k} \dots, y_{i, C}^{K}) \in \mathbb{R}^{C\times K}$ to denote all $C$ types of crimes that occurred during the past $K$ slots' observations. Following the general settings of previous crime forecasting works ~\cite{huang2018deepcrime,huang2019mist,sun2021crimeforecaster}, we set each element $y_{i, l}^{k}$ to 1 if  crime type $l$ happens at region $i$ in time slot $t$, and 0 otherwise. 
\end{myDef}

To utilize the advanced graph neural network on crime forecasting, we define a region graph with a weight matrix describing the relationships between region nodes. Also, we introduce Homophily Ratio~\cite{zhu2020beyond} of a graph with node labels which is related to our model design.
\begin{myDef}
    \textbf{Region Graph.} A region graph is formulated as $\mathcal{G} = (\mathcal{V}, \mathcal{E}, \mathcal{A}_r)$, where $\mathcal{V}$ is a set of region nodes with $|\mathcal{V}|=N$ where $N$ is the number of region nodes in the graph, and $\mathcal{E}$ is a set of directional edges between region nodes, $\mathcal{A}_r$ is the weight matrix in Def~\ref{def:adj}.
\end{myDef}
\begin{myDef}
\label{def:adj}
    \textbf{Weight Matrix.} Weight matrix is a representation of a directed graph, denoted as $\mathcal{A}_r \in \mathbb{R}^{N \times N}$ with $\mathcal{A}_{r}(i,j) = w > 0$ if $(v_i,v_j) \in \mathcal{E}$ and $\mathcal{A}_{r}(i,j) = 0$ if $(v_i,v_j) \notin \mathcal{E}$. The $\mathcal{A}_{r}(i,j)$ reflects the strength of influence from region $i$ to region $j$. 
\end{myDef}
\begin{myDef}
    \label{label:hr}
    \textbf{Homophily Ratio.} For a given graph $\mathcal{G} = (\mathcal{V}, \mathcal{E}, A)$ with node label $\mathcal{Y}$, \citet{zhu2020beyond} defines the graph's homophily ratio $\mathcal{H}(\mathcal{G},\mathcal{Y})$ to represent the probability that neighboring nodes share the same label as follows:
\end{myDef}
$$\mathcal{H}(\mathcal{G},\mathcal{Y}) = \frac{1}{|\mathcal{V}|} \sum_{v \in \mathcal{V}} \frac{|\{ u: u\in \mathcal{N}_v \land y_u =y_v\}|}{|\mathcal{N}_v|}$$
where $\mathcal{N}_v$ denotes the neighboring nodes of region node $v$ and $y_v$ is the label of node $v$.

\para{\sysname \space Goal.} Given the historical crime records across all regions from time slot $t_1$ to $t_{K}$: $\{\mathcal{Y}^1,\mathcal{Y}^2,\dots,\mathcal{Y}^k, \dots,\mathcal{Y}^K\}$, and the predefined region graph $\hat{\mathcal{G}}$, we aim to learn a function $p(\cdot)$ which can learn the adaptive region graph structure $\mathcal{G}$ and finally forecast the crime occurrences $\mathcal{Y}^{K+1}$ of all crime categories for all regions at time $t_{K+1}$, which can be formulated as follow:
$$
    \{\mathcal{Y}^1,\mathcal{Y}^2,\dots,\mathcal{Y}^k, \dots,\mathcal{Y}^K, \hat{\mathcal{G}}\} \xrightarrow{p(\cdot)} \{\mathcal{Y}^{K+1},\mathcal{G}\},
$$
\noindent where $\mathcal{Y}^k \in \mathbb{R}^{N\times C}$ denotes the crime records across all $N$ regions and $C$ crime categories in time slot $t_k$.

\section{\sysname ~Model}
\label{sec:method}
\subsection{Framework Overview}
\begin{figure*}[t]
    \centering
    \includegraphics[width=0.9\linewidth]{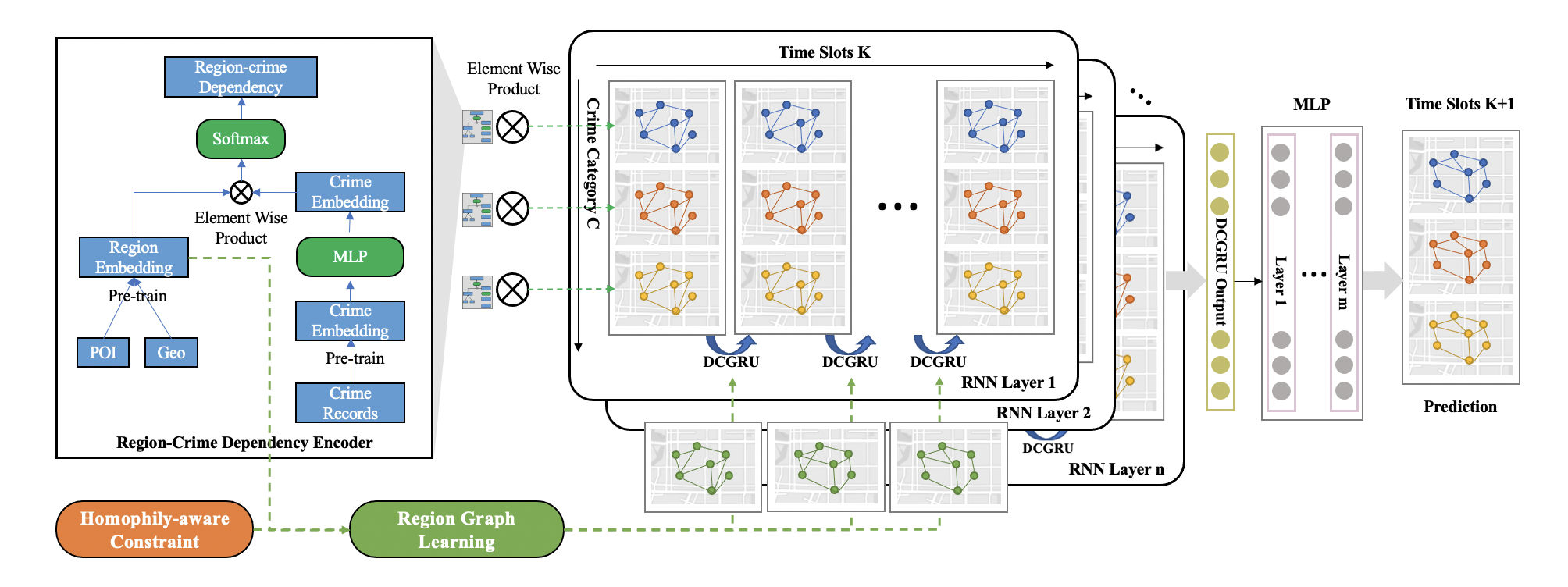}
    \caption{\sysname \space Framework.}
    \label{fig:framework}
\end{figure*}

We first explain the general framework of our model. As illustrated in Figure~\ref{fig:framework}, \sysname \space at its highest level consists of the homophily-aware graph learning layer, the diffusion graph convolution module, GRU-based temporal module, and the MLP-based decoder module. To discover hidden associations among region nodes, a graph learning layer with homophily-aware constraint learns a weight matrix, which is used as an input to the diffusion convolution modules. Diffusion convolution modules are interwoven with GRU networks to capture temporal dependencies. To obtain final predictions, the MLP-based decoder module projects the hidden features to the desired output dimension. Each core component of our model is discussed in turn as follows.

\subsection{Region Graph Learning}
A well-defined graph structure that is suitable for crime forecasting is essential for graph-based methods. 
As we illustrated in Figure~\ref{fig:case}, a pre-defined graph cannot fully capture the real connectivity in terms of crime patterns, thus affecting the final performance of crime forecasting. Therefore, we aim to learn the graph adaptively instead of using a fixed graph to reflect the crime patterns shared by regions. 

Besides, with crime forecasting, we often assume that the change of a region’s security condition indicates the change of another region's security condition~\cite{morenoff1997violent}, which can be dubbed as crime flow. Hence, we suppose the connections between region nodes in our crime-specific graph are uni-directional (i.e., if $\mathcal{A}_{r}(i,j)>0$, $\mathcal{A}_{r}(j,i)$ must be zero), which leads to better model performance in practice. 
Rather than refining the graph structure directly during the training process, we achieve the learning of the graph by updating the node embeddings adaptively and then change the graph structure accordingly.
Similar to \cite{wu2020connecting}, we design the graph learning layer based on two region embeddings: source region embedding $\mathbf{E_s}\in \mathbb{R}^{N\times D}$ and target region embedding $\mathbf{E_t}\in \mathbb{R}^{N\times D}$, where $N$ is the number of regions and $D$ is the dimension of the embedding space. We leverage pairwise region node similarity to compute the uni-directional adaptive weight matrix $\mathcal{A}_r$ as follows:
\begin{align}
    \mathbf{Z_s} &= \operatorname{tanh}(\alpha \mathbf{E_s} \mathbf{\Theta_1})\\
    \mathbf{Z_t} &= \operatorname{tanh}(\alpha \mathbf{E_t} \mathbf{\Theta_2})\\
    \mathcal{A}_r&=\operatorname{ReLU}(\operatorname{tanh}(\alpha(\mathbf{Z_s}\mathbf{Z_t}^T-\mathbf{Z_t}\mathbf{Z_s}^T)))
\end{align}
where $\alpha$ is the hyper-parameter to control the saturation rate of the hyperbolic tangent function and $\mathbf{\Theta}_1$ and $\mathbf{\Theta}_2$ denote linear transformation weights. The $\operatorname{ReLU}$ activation function leads to the asymmetric property of matrix $\mathcal{A}_r$. To ensure graph sparsity, we connect each node with the top $k$ nodes with the largest similarity. The weights between disconnected nodes are set to be zero.

To incorporate the geographical proximity between regions and have meaningful initial graph for faster convergence, we learn the embeddings of region nodes via a non-linear transformation from their pre-trained embeddings on graphs defined by geospatial and POI proximity. Particularly, 
we utilize Node2Vec~\cite{grover2016node2vec} (a classical graph embedding method) to pre-train the region embedding $\mathbf{E_{pre}}$ based on the distance-based graph and the POI similarity-based graph respectively and concatenate them together with a Multilayer Perceptron (MLP). 

\subsection{Homophily-aware Constraint}
\label{sec:homo}
The remaining question is that what factors we should consider to learn a good region graph in addition to preserving node similarity and controlling sparsity~\cite{wu2020connecting,wu2019graph}. 
Following the basic homophily assumption of graph neural network~\cite{zhu2020beyond}, we aim at learning a region graph where neighboring region nodes share similar crime patterns. Therefore, we propose a homophily constraint to the learning process for explicit heterophily reduction (i.e., reduce the probability that neighboring region nodes share totally different crime patterns).

Since the original definition of homophily ratio (see Def. ~\ref{label:hr}) is defined on static graphs for node classification under the semi-supervised setting, we need to extend it to fit our case. For our crime forecasting problem, historical crime observations in region $r_i$ and time slot $t_k$ across different crime categories $y_{i,1}^{k}, \dots, y_{i,C}^{k}$ have the additional temporal dimension $k$ and the edges are defined by the weight matrix $\mathcal{A}_{r}$ with element $\mathcal{A}_r(u, v) \in [0,1]$ representing the edge weight between region $u$ and region $v$. Therefore, given time slot $t_k$ and crime category $l$, we extend the definition of homophily ratio $\mathcal{H}(\mathcal{A}_r, \mathcal{Y}_l^{k}, l)$ as:

\begin{equation}
\begin{aligned}
\mathcal{H}(\mathcal{A}_r,\mathcal{Y}_l^{k}, l) =  \frac{1}{|\mathcal{V}|} \sum_{v \in \mathcal{V}} \frac{\sum_{u \in N(v), y_{u,l}^k=y_{v,l}^k} \mathcal{A}_r(u, v)}{\sum_{u \in N(v)} \mathcal{A}_r(u, v)}
\end{aligned}
\end{equation}
 
The extended homophily ratio is the sum of edge weight where connected region nodes share the same crime label divided by the sum of the edge weight considering whole neighbor nodes.
Intuitively, the extended homophily ratio measures the probability that neighbor region nodes share similar crime patterns, thus optimizing the homophily ratio to 1 (the maximum of homophily ratio) of our graph would boost the effectiveness of information propagation of graph convolution in Section~\ref{sec:diffusion} and finally contribute to crime forecasting.
Therefore, with definition of the extended homophily ratio, we formally define loss $\mathcal{L}_{homo}$ to regularize homophily of the learnt graph from time $t_1$ to $t_K$ as follows:
\begin{equation}
\label{equ:homo_constraint}
\begin{aligned}
    \mathcal{L}_{homo} = \Sigma_{k=1}^{K}\Sigma_{l=1}^{C} [\mathcal{H}(\mathcal{A}_r,\mathcal{Y}_l^{k}, l) - 1]^{2}
\end{aligned}
\end{equation}
where $C$ denotes the number of crime categories and $\mathcal{Y}_l^{k}$ is the records of crime category $l$ at time slot $t_k$. 

\subsection{Graph Diffusion Convolution}
\label{sec:diffusion}
In the graph diffusion convolution module, we propose a direction-aware diffusion convolution layer to simulate real-life crime diffusion and capture crime dependencies between regions. We concern the propagation of crime patterns between regions as a diffusion process from one region's first order neighbors to its $M$-th order neighbors or in the reversed way. To learn such diffusion pattern, similar to~\cite{li2018dcrnn_traffic}, we characterize graph diffusion on $\mathcal{G}$ with the $m$-th order transition matrices $\mathbf{S}_m^O, \mathbf{S}_m^I$ on out-degree and in-degree basis respectively:
\begin{align}
    \mathbf{S}_m^O &= (\mathbf{D}_O^{-1}\mathcal{A}_r)^m \\
    \mathbf{S}_m^I &= (\mathbf{D}_I^{-1}\mathcal{A}_r^{\top})^m
\end{align}
where $\mathbf{D}_O$ and $\mathbf{D}_I$ are diagonal matrices of nodes out-degree and in-degree respectively. Different from~\cite{li2018dcrnn_traffic}, the weight matrix $\mathcal{A}_r$ is calculated with unidirectional constraints. Thus the diffusion matrices $\mathbf{S}_m^O, \mathbf{S}_m^I$ will be sparse and also follow the unidirectional constraint during the $m$-step diffusion.

Given the above transition matrices, we then define a direction-aware diffusion convolution function  that transform the input $\mathbf{X} \in \mathbb{R}^{N\times H}$ from all nodes into (hidden) outputs via the diffusion process. 
Here if $\mathbf{X}$ is the input, $H$ equals to number of crime categories $C$, otherwise it is the dimension of the hidden states. 
Specifically, the direction-aware diffusion convolution function $f_*(\mathbf{X}; \mathcal{G}, \mathbf{\Theta}, \mathbf{D_W})$ aggregates the input $\mathbf{X}$ on graph $\mathcal{G}(\mathcal{V}, \mathcal{E}, \mathcal{A}_r)$ with parameters $\mathbf{\Theta}$ and $\mathbf{D_W}$ in the following way:
\begin{align}
\label{equ:diffusion}
    f_*(\mathbf{X}; \mathcal{G}, \mathbf{\Theta}, \mathbf{D_W}) = \sum_{m=0}^{M}(\mathbf{S}_m^O \mathbf{X} \mathbf{\Theta_{:,:,m,1}} + \mathbf{D_W} \mathbf{S}_m^I \mathbf{X} \mathbf{\Theta_{:,:,m,2}})
\end{align}
where $M$ is the total diffusion step, $\mathbf{\Theta} \in \mathbb{R}^{H \times H' \times (M+1) \times 2}$ ($H'$ is the output dimension) is the filter parameter. $\mathbf{D_W}$ is a diagonal matrix designed for measuring the direction preference of each node. 
Specifically, the $i_{th}$ diagonal value of $\mathbf{D_W}$ denotes the the preference of region $i$ on in-going diffusion. 
A high value means the region prefers in-going diffusion to out-going diffusion, i.e., the region is more likely to be affected by other regions than to affect other regions.
Via this direction-aware diffusion convolution function, each node (region) would learn its own way to perform the aggregation of crime patterns from neighboring nodes and produce meaningful (hidden) outputs for further computation. 

\subsection{Temporal Module}
For temporal crime dynamics, we use RNN with DCGRU (Diffusion Convolutional Gated Recurrent) units~\cite{li2018dcrnn_traffic} in \sysname \space to capture the complex crime connectivity across historical time slots (from $t_1$ to $t_K$). In DCGRU, the original linear transformation in GRU is replaced by the direction-aware diffusion convolution in Equation~\ref{equ:diffusion}, which can incorporate global information from the whole graph and enable the learning process of a node to be based on not only its previous state but also its neighbors' previous state with similar crime patterns. We formulate the updating functions for hidden state $h^{t}$ at $t$-th time step of input signal sequence in our DCGRU-revised RNN encoder as follows:
\begin{align}
\nonumber r^t  &= \sigma(f_* ([x^t, h^{t-1}]; \mathcal{G}, \mathbf{\Theta_r}, \mathbf{D_W}) + \mathbf{b_r})\\
\nonumber u^t  &= \sigma(f_* ([x^t, h^{t-1}]; \mathcal{G}, \mathbf{\Theta_u}, \mathbf{D_W}) + \mathbf{b_u})\\\nonumber
    \nonumber c^t  & = tanh(f_* ([x^t, r^t\odot h^{t-1}]; \mathcal{G}, \mathbf{\Theta_c}, \mathbf{D_W}) + \mathbf{b_c})\\
h^t &= u^t \odot h^t + (1-u^t)\odot c^t
\end{align}
where $x^t, h^t$ are the input and output of the DCGRU cell at time $t$, and $r^t, u^t, c^t$ are the reset gate, update gate and cell state respectively. The parameters $\mathbf{\Theta_r}, \mathbf{\Theta_u}$ and $\mathbf{\Theta_c}$ are filter parameters of diffusion convolution and $\mathbf{b_r}, \mathbf{b_u}$ and $\mathbf{b_c}$ are the bias terms. All gates share the same direction weight $\mathbf{D_W}$, indicating the proportional intensity of reverse diffusion is consistent for the same region. The parameter $\sigma$ denotes the sigmoid function and $\odot$ denotes the Hadamard Product. 
From the encoder-decoder perspective, as shown in Figure~\ref{fig:framework}, \sysname's encoder consists of multiple stacked layers of RNNs with DCGRU units, which capture temporal transitions across regions and time. The encoder's outputs, i.e., the final hidden states, are delivered to the decoder for future crime forecasting.

\subsection{Region-crime Dependency Encoder}
As depicted in the left part of Figure~\ref{fig:framework}, in order to capture the underlying dependencies between regions and crime categories, we introduce the crime embedding $\mathbf{E}_c \in \mathbb{R}^{C\times D}$ for all $C$ crime categories and compute the inter region-crime dependency matrix $\mathbf{W_{inter}}\in \mathbb{R}^{N\times C}$ by calculating an element-wise product between region embedding and crime embedding with a transition matrix. 
Each element of the inter dependency matrix $\mathbf{W_{inter}}(i,l)$ represents the dependency between region $i$ and crime category $l$. After normalizing the dependency weight via a softmax function, we perform an element-wise product between dependency weight matrix $\mathbf{W_{inter}}$ and the input crime records $\mathcal{Y}^k\in \mathbb{R}^{N\times C}$ across all regions and categories in $k$-th time slot to generate a weighted input $\mathbf{W_{inter}} \odot \mathcal{Y}^k\in \mathbb{R}^{N\times C}$ which is then fed into the encoder (i.e., RNN with DCGRU units) of \sysname. 

To incorporate the prior knowledge of crime patterns, we flatten the training records across all regions and time slots for each crime category and use PCA~\cite{martinez2001pca} for dimension reduction to initialize the crime embedding $\mathbf{E}_c \in \mathbb{R}^{C\times D}$.

\subsection{Crime Forecasting and Model Inference}
In general, we employ the Encoder-Decoder architecture~\cite{cho2014properties} for the crime forecasting. 
After encoding the temporal crime dynamics by the encoder, \sysname \space utilizes a Multilayer Perceptron (MLP) based decoder to map the encoded hidden states to the outputs for crime forecasting in a non-linear way. We formulate the decoder (i.e., MLP with diffusion convolutional layers) as follows.
\begin{align}
\nonumber \psi_1 &= \text{ReLU}(f_* (h; \mathcal{G}, \mathbf{W}_1,\mathbf{D_W})+ b_1)\\
\nonumber ...\\
\psi_P &= f_*(\psi_{P-1}; \mathcal{G}, \mathbf{W}_P, \mathbf{D_W})+ b_P)\\
\nonumber y &= \sigma(f_*(\psi_P; \mathcal{G}, \mathbf{W}^{\prime}, \mathbf{D_W})+b^{\prime})
\end{align}
where $h$ is the input of the decoder and is retrieved from the output of temporal module, i.e., $h^t$. The output $y$ is a matrix denoting the crime occurrence probabilities of all regions for all categories at the prediction time $t_{K+1}$. During the training process, we use binary cross entropy as the major learning objective:
\begin{equation}
\begin{aligned}
\label{eq:loss_crime}
\mathcal{L}_{crime} = -\sum_{i\in \{1,\dots,R\}, \atop l\in \{1,\dots,C\}} y_{i,l}\log \hat{y}_{i,l} + (1-y_{i,l})\log(1-\hat{y}_{i,l}),
\end{aligned}
\end{equation}
where $y_{i,l}$ and $\hat{y}_{i,l}$ represent the empirical and estimated probability of the $l$-th crime category happening at region $i$ at the time $t_{K+1}$, respectively. 

As mentioned in Section~\ref{sec:homo}, we propose a homphily-aware constraint loss during the graph learning process to conform to the homophily assumption of GNN. A natural approach is to optimize linear combination of the corresponding loss functions, which is formulated as follows:
\begin{equation}
\begin{aligned}
\label{eq:loss}
\mathcal{L}_{total} = \mathcal{L}_{crime} + \lambda \mathcal{L}_{homo},
\end{aligned}
\end{equation}
where $\lambda$ is the trade-off parameter.  We minimize the joint loss function by using the Adam optimizer~\cite{kingma2014adam} with learning rate decay strategy to learn the parameters of \sysname .
\section{Experiments}
\label{sec:exp}
We conduct our experiments on two real-world datasets and elaborate the following points in this section. First, we explore \sysname's performance as compared to the state-of-the-art spatial and temporal graph neural networks and crime forecasting models, the performance in different crime categories, and how the key components affect \sysname's performance in the joint framework in Section \ref{subsec.perform}. Then, we focus on model explainability to gain meaningful insights from the learned region graph in Section \ref{subsec.explain}.

\subsection{Data and Experimental Setting}

\subsubsection{Data Description and Training Configuration}
We evaluated \sysname \space on two real-world benchmarks in Chicago and Los Angeles by CrimeForecaster~\cite{sun2021crimeforecaster}. In our experiments, we use the same ``train-validation-test'' setting as the previous work ~\cite{sun2021crimeforecaster, huang2019mist}. We chronologically split the dataset as 6.5 months for training, 0.5 months for the validation, and 1 month for testing. For the vital hyperparameters in \sysname, we use two stacked layers of RNNs. Within each RNN layer, we set 64 as the size of the hidden dimension. Moreover, we set the subgraph size of the sparsity operation as 50 and the saturation rate as 3. For the learning objective, we fix the trade-off parameter $\lambda$ as $0.01$, similar to the common practice of other regularizers. 

\subsubsection{Evaluation Metric}
Since the crime forecasting problem can be viewed as a multi-label classification problem, we utilize Micro-F1~\cite{grover2016node2vec} and Macro-F1~\cite{lin2020healthwalks} as general metrics to evaluate prediction performance across all crime categories, similar to CrimeForecaster~\cite{sun2021crimeforecaster} and MiST~\cite{huang2019mist}. 

\subsubsection{Baseline}
\label{subsec.baseline}
We choose three categories of baselines.

\emph{Traditional Methods.} We use three types of traditional methods, including time series forecasting model ARIMA~\cite{contreras2003arima}, classical machine learning methods Epsilon-Support Vector Regression (SVR)~\cite{chang2011libsvm},  Decision Tree~\cite{safavian1991survey} and Random Forest~\cite{verikas2011mining}, and traditional neural networks Multi-layer Perceptron classifier (MLP)~\cite{covington2016deep}, Long Short-term Memory (LSTM)~\cite{gers1999learning} and Gated Recurrent Unit (GRU)~\cite{chung2014empirical}.

\emph{Spatial-temporal graph neural network} (i.e., MTGNN and Graph WaveNet). Among the spatial-temporal graph neural network models in Section \ref{subsec.stgnn}, we select two methods designed for multivariate time-series forecasting which are generally used as baselines in previous works for comparison. Graph WaveNet (GW)~\cite{wu2019graph} is a spatial-temporal graph neural network that combines diffusion convolutions with 1D dilated convolutions. MTGNN~\cite{wu2020connecting} is the state-of-the-art spatial-temporal graph neural network in previous works, which integrates the adaptive graph structure, mix-hop graph convolution layers, and dilated temporal convolution layers.

\emph{Spatial-temporal learning models for crime forecasting }(i.e., MiST and CrimeForecaster). MiST~\cite{huang2019mist} learns both the inter-region temporal and spatial correlations.  
To the best of our knowledge, CrimeForecaster (CF)~\cite{sun2021crimeforecaster} is the most recent crime forecasting model, which is an end-to-end framework to model the dependencies between regions by geographical neighborhoods instead of grid partitions and captures both spatial and temporal dependencies using diffusion convolution layer and Gated Recurrent Network.

\subsection{Performance Evaluation}
\label{subsec.perform}

\begin{table*}[htbp]
    \tiny
  \centering
  \caption{Performance comparison with the state-of-the-art baselines on crime forecasting.}
    \begin{tabular}{c|c|c|cccccccccccc}
    \toprule
    Data  & Month & Metric & ARIMA & LSTM  & GRU   & MLP   & DT    & SVR   & RF    & MiST  & GW    & MTGNN & CF    & \textbf{HAGEN} \\
    \midrule
    \midrule
    \multirow{10}[10]{*}{CHI} & \multirow{2}[2]{*}{8} & Micro-F1 & 0.4867  & 0.5032  & 0.5047  & 0.5443  & 0.6226  & 0.5822  & 0.6860  & 0.6719  & 0.6593  & 0.6876  & 0.7052  & \textbf{0.7209 } \\
          &       & Macro-F1 & 0.4205  & 0.4371  & 0.4339  & 0.4515  & 0.4324  & 0.3424  & 0.3423  & 0.6176  & 0.6201  & 0.6516  & 0.6636  & \textbf{0.6791 } \\
\cmidrule{2-15}          & \multirow{2}[2]{*}{9} & Micro-F1 & 0.4836  & 0.4979  & 0.4818  & 0.5077  & 0.6252  & 0.6064  & 0.6967  & 0.6892  & 0.6512  & 0.6885  & 0.6929  & \textbf{0.7211 } \\
          &       & Macro-F1 & 0.3979  & 0.4155  & 0.4206  & 0.4326  & 0.4404  & 0.3288  & 0.3587  & 0.6351  & 0.6111  & 0.6524  & 0.6491  & \textbf{0.6779 } \\
\cmidrule{2-15}          & \multirow{2}[2]{*}{10} & Micro-F1 & 0.4757  & 0.4834  & 0.4886  & 0.4240  & 0.6166  & 0.6395  & 0.6933  & 0.6692  & 0.6456  & 0.6832  & 0.6931  & \textbf{0.7129 } \\
          &       & Macro-F1 & 0.3881  & 0.4031  & 0.4125  & 0.3944  & 0.4396  & 0.3352  & 0.3624  & 0.6211  & 0.6068  & 0.6494  & 0.6506  & \textbf{0.6738 } \\
\cmidrule{2-15}          & \multirow{2}[2]{*}{11} & Micro-F1 & 0.4520  & 0.4495  & 0.4657  & 0.5449  & 0.6108  & 0.5935  & 0.6833  & 0.6766  & 0.6238  & 0.6539  & 0.6774  & \textbf{0.6946 } \\
          &       & Macro-F1 & 0.3667  & 0.3988  & 0.4059  & 0.4256  & 0.4300  & 0.3296  & 0.3523  & 0.6262  & 0.5849  & 0.6153  & 0.6356  & \textbf{0.6528 } \\
\cmidrule{2-15}          & \multirow{2}[2]{*}{12} & Micro-F1 & 0.4528  & 0.4891  & 0.4655  & 0.4957  & 0.6133  & 0.6261  & 0.6795  & 0.6753  & 0.5933  & 0.6480  & 0.6773  & \textbf{0.6907 } \\
          &       & Macro-F1 & 0.3697  & 0.4041  & 0.4034  & 0.4146  & 0.4247  & 0.3325  & 0.3540  & 0.6138  & 0.5581  & 0.6064  & 0.6379  & \textbf{0.6467 } \\
    \midrule
    \midrule
    \multirow{10}[10]{*}{LA} & \multirow{2}[2]{*}{8} & Micro-F1 & 0.3711  & 0.4159  & 0.3931  & 0.4075  & 0.5072  & 0.4569  & 0.5798  & 0.5991  & 0.5699  & 0.5561  & 0.6038  & \textbf{0.6216 } \\
          &       & Macro-F1 & 0.3036  & 0.3171  & 0.3285  & 0.3236  & 0.3013  & 0.2119  & 0.2054  & 0.4920  & 0.5079  & 0.4871  & 0.5359  & \textbf{0.5509 } \\
\cmidrule{2-15}          & \multirow{2}[2]{*}{9} & Micro-F1 & 0.3668  & 0.4005  & 0.4357  & 0.3636  & 0.5104  & 0.4748  & 0.5849  & 0.5998  & 0.5719  & 0.5792  & 0.6035  & \textbf{0.6210 } \\
          &       & Macro-F1 & 0.2959  & 0.3118  & 0.3160  & 0.3162  & 0.3019  & 0.2005  & 0.2097  & 0.4877  & 0.4973  & 0.5042  & 0.5180  & \textbf{0.5443 } \\
\cmidrule{2-15}          & \multirow{2}[2]{*}{10} & Micro-F1 & 0.3722  & 0.4010  & 0.3994  & 0.3728  & 0.5147  & 0.5302  & 0.5742  & 0.5956  & 0.5611  & 0.5590  & 0.5886  & \textbf{0.6165 } \\
          &       & Macro-F1 & 0.3010  & 0.3174  & 0.3197  & 0.3225  & 0.3114  & 0.1992  & 0.4492  & 0.4754  & 0.4891  & 0.4915  & 0.5028  & \textbf{0.5455 } \\
\cmidrule{2-15}          & \multirow{2}[2]{*}{11} & Micro-F1 & 0.3800  & 0.4688  & 0.4643  & 0.4639  & 0.5058  & 0.4899  & 0.5790  & 0.5393  & 0.5711  & 0.5736  & 0.5922  & \textbf{0.6190 } \\
          &       & Macro-F1 & 0.3090  & 0.3774  & 0.3890  & 0.3797  & 0.2975  & 0.2048  & 0.4519  & 0.4999  & 0.5002  & 0.4994  & 0.5083  & \textbf{0.5373 } \\
\cmidrule{2-15}          & \multirow{2}[2]{*}{12} & Micro-F1 & 0.3730  & 0.4482  & 0.4543  & 0.4481  & 0.5103  & 0.5294  & 0.5685  & 0.5343  & 0.5570  & 0.5663  & 0.5841  & \textbf{0.6113 } \\
          &       & Macro-F1 & 0.3010  & 0.3849  & 0.3662  & 0.3856  & 0.3033  & 0.2084  & 0.4401  & 0.4944  & 0.4843  & 0.4954  & 0.4921  & \textbf{0.5250 } \\
    \bottomrule
    \end{tabular}
  \label{tab:performance}
\end{table*}

\subsubsection{Overall Performance} 
\label{overallperform}
Table~\ref{tab:performance} shows the crime forecasting accuracy in both Chicago and Los Angeles dataset. We evaluate the performance of crime forecasting in terms of Macro-F1 and Micro-F1 for all methods. We summarize the following key observations from Table~\ref{tab:performance}:

\emph{Advantage of graph-based model.}
For the three types of baseline models stated in Section~\ref{subsec.baseline}, graph-based methods generally outperform non-graph-based ones. We share the same conclusion with CrimeForcaster that graphs can better capture spatial relation between regions, compared with both grid-based models (i.e., MiST) and traditional methods.

\emph{Advantage of graph learning with the homophily-aware constraint.}
\sysname \space consistently outperforms the state-of-the-art CrimeForecaster in 5 testing months on both datasets in Micro-F1 and Macro-F1, which showcases the effectiveness of our adaptively-learned graph. In addition, \sysname \space outperforms other models with adaptive graph but without homophily-aware constraint (i.e., MTGNN and Graph WaveNet), verifying the importance of incorporating the homophily-aware constraint when learning the region graph.
\begin{figure}[htbp]
    \centering
    \includegraphics[width=1.0\linewidth]{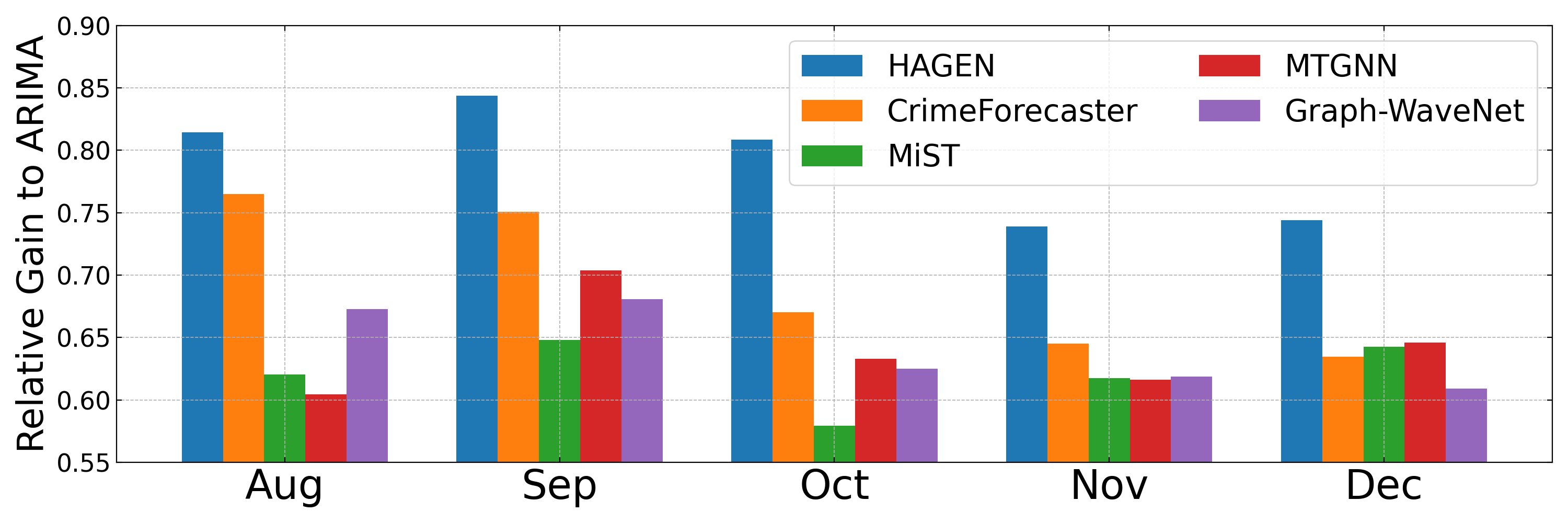}
    \caption{Relative Macro-F1 gain of models over ARIMA  on LA dataset.}
    \label{fig:overall_la}
\end{figure}

\sysname \space performs significantly better than all other learning models. In particular, the most notable advancement occurs in crime forecasting of Los Angeles in October, November, and December. \sysname \space reaches a 5.50\% relative improvement over the state-of-the-art (i.e., CrimeForecaster) in terms of Micro-F1 and an 8.25\% relative improvement in terms of Macro-F1 in October.
We take ARIMA as the base model and display the relative gain of competitive models over it in Figure~\ref{fig:overall_la}. The improvement of CrimeForecaster as compared to the previous state-of-the-art, MiST, is less than that of \sysname \space over CrimeForecaster. 

\subsubsection{Performance Comparison Across Crime Categories}
\begin{figure}[htbp]
    \centering
    \includegraphics[width=1.0\linewidth]{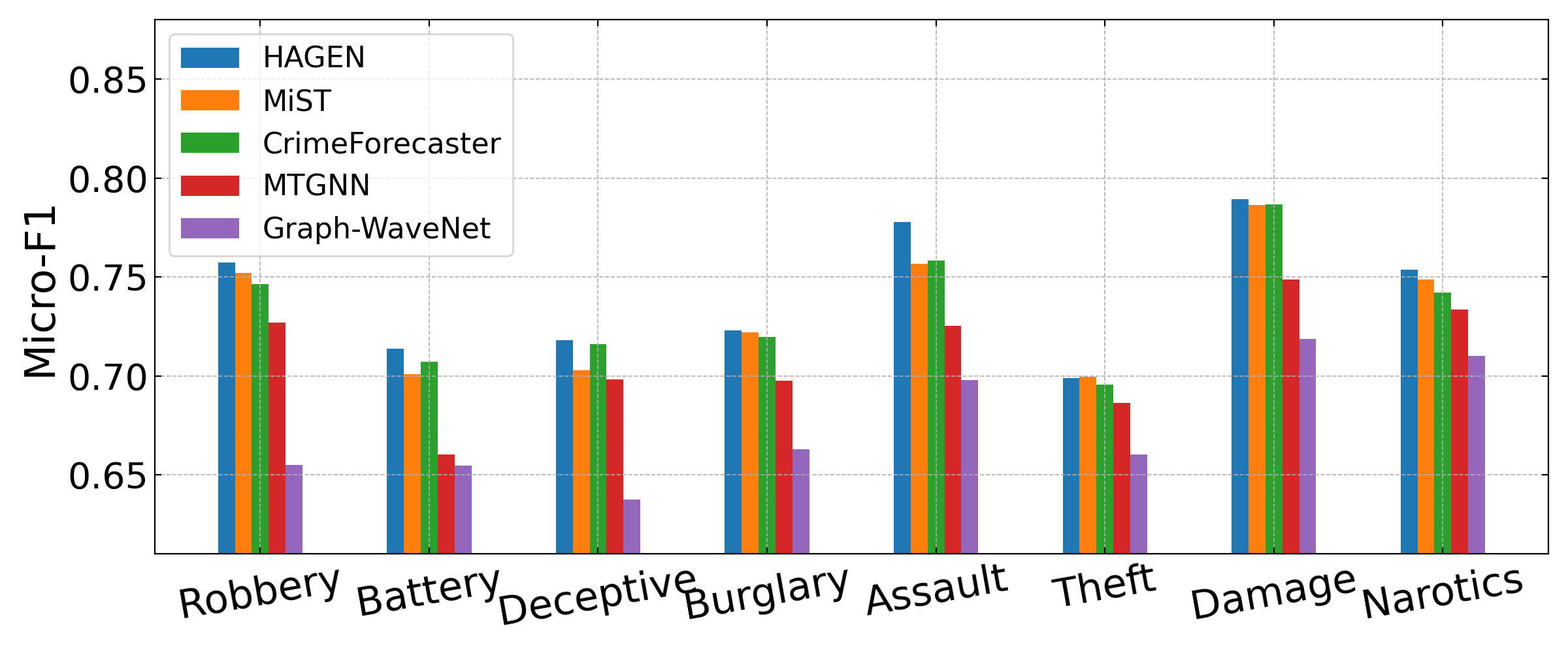}
    \caption{Micro-F1 for individual crime categories on Chicago dataset.}
    \label{fig:cat_la}
\end{figure}

We also evaluate \sysname’s performance across different crime categories with those of a selected competitive baselines (i.e., CrimeForecaster, MTGNN, and Graph WaveNet) on the Chicago dataset. As shown in Figure~\ref{fig:cat_la}, \sysname \space achieves almost consistent performance gain in Micro-F1 score across various crime categories compared to the competitors, showing the effectiveness of our model. 

\subsubsection{Ablation Study}
\begin{figure}[htbp]
\centering
\subfigure[Micro-F1]{
\label{fig:ablation_micro}
\includegraphics[width=0.42\linewidth]{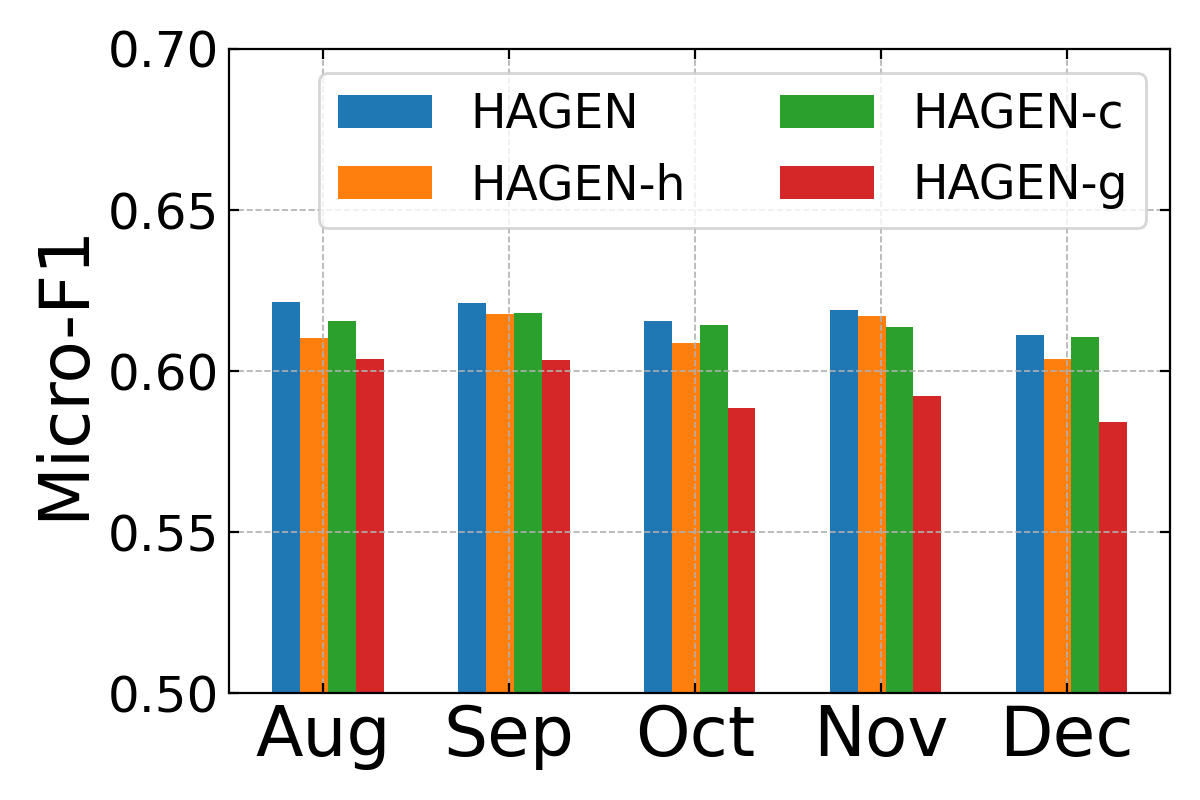}}
\subfigure[Macro-F1]{
\label{fig:ablation_macro}
\includegraphics[width=0.42\linewidth]{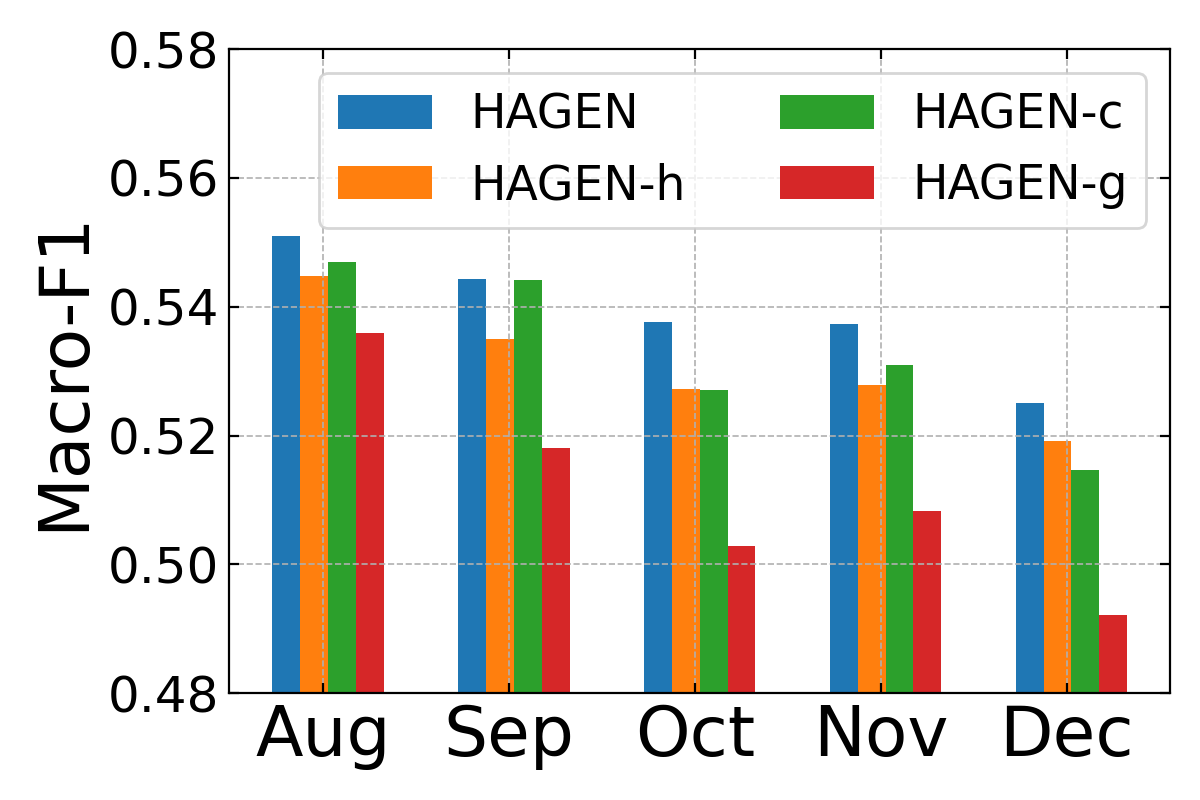}}
\caption{Evaluation on the ablated variants of \sysname.}
\end{figure}

We evaluate how each key component contributes to our framework. We consider three degrade variants: \sysname-h, \sysname-c, \sysname-g, which removes the homophily-aware constraint, region-crime dependency and graph learning layer from \sysname\space respectively. We compare the complete \sysname \space with variants in Figure~\ref{fig:ablation_micro} and \ref{fig:ablation_macro}. We observe that taking out each component in \sysname \space will lead to a performance drop, which indicates the importance of each component and justifies our model design. Specifically, \sysname-g incurs the worst performance, suggesting the graph learning layer to be the most impactful component of \sysname.

\subsection{Model Explainability}
\label{subsec.explain}

\begin{figure}[htbp]
    \centering
    \includegraphics[width=0.4\linewidth]{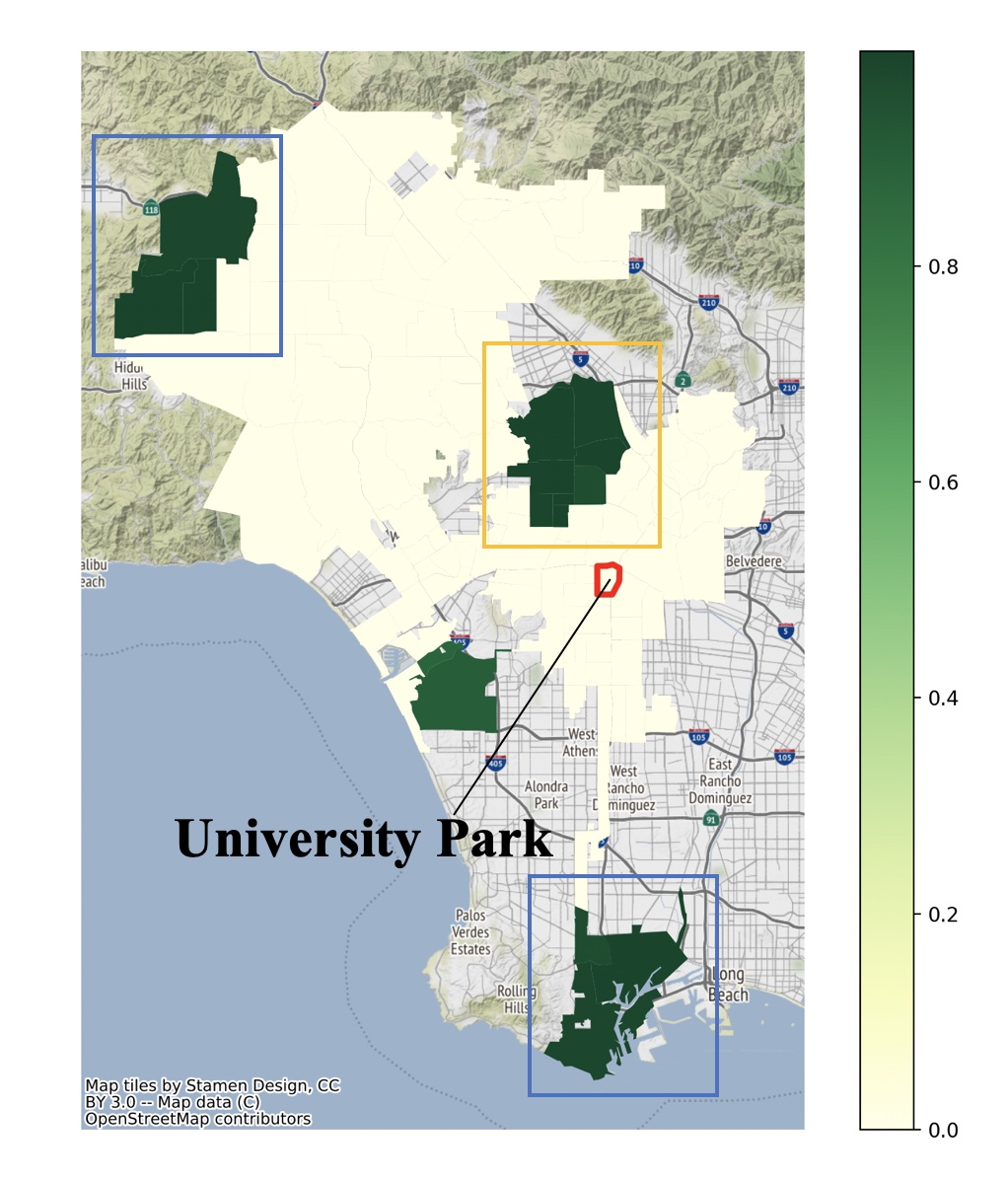}
    \caption{Visualization of adaptively learned neighbors of University Park.}
    \label{fig:glmap}
\end{figure}

To have a better understanding of what \sysname \space acquired in the graph learning process, we analyze the hidden relationship revealed by the learned weight matrix by comparing the neighboring nodes of some popular regions in the pre-defined graph and in the learned graph. We take the ``University Park'' region of Los Angeles as an example in Figure \ref{fig:glmap} to illustrate the pattern. If we consider the graph defined by geographical distance, University Park is closely connected with regions next to it like Downtown. However, a major university resides in University Park, which deploys strong police force, and hence it should have a very different pattern from its adjacent neighbors in term of crime events.

In contrast, the adaptive region graph is more successful in capturing the crime-related proximity as depicted in the new ``neighbors'' of University Park in the graph adaptively learned by \sysname \space in Figure \ref{fig:glmap}. These learned neighbors can be categorized into two classes: remote and less-populated regions like Chatsworth Reservoir and Harbor City (framed in blue) and more secure regions like Hancock Park (framed in yellow). These neighborhoods better resemble University Park with respect to crime occurrence.

\section{Related Works}
\label{sec:relatedworks}

\subsection{Forecasting Model for Crime Forecasting}

As a prediction problem of the sequential data, it is natural to utilize models like ARIMA and LSTM~\cite{mei2019predictability,safat2021empirical} in crime forecasting. To further capture the spatial dependency of crime events as well as temporal dependency, Huang et al. partition the region into synthetic units and combined CNN-based models and RNN-based models in DeepCrime~\cite{huang2018deepcrime} and MiST~\cite{huang2019mist}. The state-of-the-art CrimeForester~\cite{sun2021crimeforecaster} shows that distance-based region graph structure can better capture crime correlations. However, pre-defined graph like distance based 
weight matrix is hard to take each case into account, unavoidably leading to cases where neighboring nodes do not share similar crime pattern, which limits the performance of GNN-based models. To handle this problem, we propose an adaptive graph structure by introducing graph learning layer into \sysname \space that continuously updates during training.

\subsection{Spatial-temporal Graph Neural Network}
\label{subsec.stgnn}

In traffic forecasting domain, many spatial-temporal graph neural network models are proposed such as DCRNN \cite{li2018dcrnn_traffic}, STGCN~\cite{yu2017spatio}, GraphWaveNet~\cite{wu2019graph} and GMAN~\cite{zheng2020gman} to model both spatial correlations and temporal dependencies. Adaptive graph learning is utilized by AGCRN~\cite{bai2020adaptive}, SLCNN~\cite{zhang2020spatio} and MTGNN~\cite{wu2020connecting} recently, and MTGNN is the state-of-the-art approach in previous papers. However, the current approaches of graph construction are heuristic with consideration of barely node similarity, graph sparsity and symmetry. Homophily, which is one of the fundemental assumptions for GNNs, is not taken into account. We proposed a homophily-aware graph convolutional recurrent network framework by explicitly introducing homophily constraint into our model to regularize the process of graph learning.

\section{Conclusion}
\label{sec:conclusion}
We present \sysname, an end-to-end graph convolutional recurrent network with a novel homophily-aware graph learning module for crime forecasting. In particular, \sysname \space uses an adaptive graph structure to capture the dependency of crime patterns between regions and incorporates direction-aware diffusion convolution layer with Gated Recurrent Network to learn spatiotemporal dynamics. The graph structure is constrained by a designed homophily-aware loss to enhance the performance of the graph neural network. We evaluate \sysname \space on two real-world benchmarks. \sysname \space consistently outperforms state-of-the-art crime forecasting model and spatiotemporal graph neural networks. 
In future work, we will improve our model and evaluate benchmarks for traffic forecasting to prove \sysname's generality for multivariate time series forecasting. Furthermore, we will explore the theoretical foundation of adaptive graph construction and how it promotes multivariate time series forecasting.
\bibstyle{aaai22}
\bibliography{refer}

\begin{thebibliography}{28}
\providecommand{\natexlab}[1]{#1}

\bibitem[{Bai et~al.(2020)Bai, Yao, Li, Wang, and Wang}]{bai2020adaptive}
Bai, L.; Yao, L.; Li, C.; Wang, X.; and Wang, C. 2020.
\newblock Adaptive Graph Convolutional Recurrent Network for Traffic
  Forecasting.
\newblock \emph{arXiv preprint arXiv:2007.02842}.

\bibitem[{Chang and Lin(2011)}]{chang2011libsvm}
Chang, C.-C.; and Lin, C.-J. 2011.
\newblock LIBSVM: A library for support vector machines.
\newblock \emph{ACM transactions on intelligent systems and technology (TIST)},
  2(3): 1--27.

\bibitem[{Chien et~al.(2021)Chien, Peng, Li, and
  Milenkovic}]{chien2021adaptive}
Chien, E.; Peng, J.; Li, P.; and Milenkovic, O. 2021.
\newblock Adaptive Universal Generalized PageRank Graph Neural Network.
\newblock In \emph{International Conference on Learning Representations.
  https://openreview. net/forum}.

\bibitem[{Cho et~al.(2014)Cho, Van~Merri{\"e}nboer, Bahdanau, and
  Bengio}]{cho2014properties}
Cho, K.; Van~Merri{\"e}nboer, B.; Bahdanau, D.; and Bengio, Y. 2014.
\newblock On the properties of neural machine translation: Encoder-decoder
  approaches.
\newblock \emph{arXiv preprint arXiv:1409.1259}.

\bibitem[{Chung et~al.(2014)Chung, Gulcehre, Cho, and
  Bengio}]{chung2014empirical}
Chung, J.; Gulcehre, C.; Cho, K.; and Bengio, Y. 2014.
\newblock Empirical evaluation of gated recurrent neural networks on sequence
  modeling.
\newblock \emph{arXiv preprint arXiv:1412.3555}.

\bibitem[{Contreras et~al.(2003)Contreras, Espinola, Nogales, and
  Conejo}]{contreras2003arima}
Contreras, J.; Espinola, R.; Nogales, F.~J.; and Conejo, A.~J. 2003.
\newblock ARIMA models to predict next-day electricity prices.
\newblock \emph{IEEE transactions on power systems}, 18(3): 1014--1020.

\bibitem[{Covington, Adams, and Sargin(2016)}]{covington2016deep}
Covington, P.; Adams, J.; and Sargin, E. 2016.
\newblock Deep neural networks for youtube recommendations.
\newblock In \emph{Proceedings of the 10th ACM conference on recommender
  systems}, 191--198.

\bibitem[{Gers, Schmidhuber, and Cummins(1999)}]{gers1999learning}
Gers, F.~A.; Schmidhuber, J.; and Cummins, F. 1999.
\newblock Learning to Forget: Continual Prediction with LSTM.
\newblock In \emph{Istituto Dalle Molle Di Studi Sull Intelligenza
  Artificiale}. IET.

\bibitem[{Grover and Leskovec(2016)}]{grover2016node2vec}
Grover, A.; and Leskovec, J. 2016.
\newblock node2vec: Scalable feature learning for networks.
\newblock In \emph{ACM SIGKDD Conference on Knowledge Discovery and Data
  Mining}, 855--864.

\bibitem[{Huang et~al.(2019)Huang, Zhang, Zhao, Wu, Yin, and
  Chawla}]{huang2019mist}
Huang, C.; Zhang, C.; Zhao, J.; Wu, X.; Yin, D.; and Chawla, N. 2019.
\newblock Mist: A multiview and multimodal spatial-temporal learning framework
  for citywide abnormal event forecasting.
\newblock In \emph{The Web Conference (WWW)}, 717--728.

\bibitem[{Huang et~al.(2018)Huang, Zhang, Zheng, and
  Chawla}]{huang2018deepcrime}
Huang, C.; Zhang, J.; Zheng, Y.; and Chawla, N.~V. 2018.
\newblock DeepCrime: Attentive hierarchical recurrent networks for crime
  prediction.
\newblock In \emph{ACM International Conference on Information and Knowledge
  Management (CIKM)}, 1423--1432.

\bibitem[{Kingma and Ba(2014)}]{kingma2014adam}
Kingma, D.~P.; and Ba, J. 2014.
\newblock Adam: A method for stochastic optimization.
\newblock \emph{arXiv preprint arXiv:1412.6980}.

\bibitem[{Li et~al.(2018)Li, Yu, Shahabi, and Liu}]{li2018dcrnn_traffic}
Li, Y.; Yu, R.; Shahabi, C.; and Liu, Y. 2018.
\newblock Diffusion Convolutional Recurrent Neural Network: Data-Driven Traffic
  Forecasting.
\newblock In \emph{The International Conference on Learning Representations}.

\bibitem[{Lin et~al.(2020)Lin, Lyu, Cao, Xu, Wei, Samet, and
  Li}]{lin2020healthwalks}
Lin, Z.; Lyu, S.; Cao, H.; Xu, F.; Wei, Y.; Samet, H.; and Li, Y. 2020.
\newblock HealthWalks: Sensing Fine-grained Individual Health Condition via
  Mobility Data.
\newblock \emph{Proceedings of the ACM on Interactive, Mobile, Wearable and
  Ubiquitous Technologies}, 4(4): 1--26.

\bibitem[{Martinez and Kak(2001)}]{martinez2001pca}
Martinez, A.~M.; and Kak, A.~C. 2001.
\newblock Pca versus lda.
\newblock \emph{IEEE transactions on pattern analysis and machine
  intelligence}, 23(2): 228--233.

\bibitem[{Mei and Li(2019)}]{mei2019predictability}
Mei, Y.; and Li, F. 2019.
\newblock Predictability comparison of three kinds of robbery crime events
  using LSTM.
\newblock In \emph{Proceedings of the 2019 2nd International Conference on Data
  Storage and Data Engineering}, 22--26.

\bibitem[{Morenoff and Sampson(1997)}]{morenoff1997violent}
Morenoff, J.~D.; and Sampson, R.~J. 1997.
\newblock Violent crime and the spatial dynamics of neighborhood transition:
  Chicago, 1970--1990.
\newblock \emph{Social forces}, 76(1): 31--64.

\bibitem[{Safat, Asghar, and Gillani(2021)}]{safat2021empirical}
Safat, W.; Asghar, S.; and Gillani, S.~A. 2021.
\newblock Empirical Analysis for Crime Prediction and Forecasting Using Machine
  Learning and Deep Learning Techniques.
\newblock \emph{IEEE Access}.

\bibitem[{Safavian and Landgrebe(1991)}]{safavian1991survey}
Safavian, S.~R.; and Landgrebe, D. 1991.
\newblock A survey of decision tree classifier methodology.
\newblock \emph{IEEE transactions on systems, man, and cybernetics}, 21(3):
  660--674.

\bibitem[{Sun et~al.(2021)Sun, Yue, Lin, Yang, Nocera, Kahn, and
  Shahabi}]{sun2021crimeforecaster}
Sun, J.; Yue, M.; Lin, Z.; Yang, X.; Nocera, L.; Kahn, G.; and Shahabi, C.
  2021.
\newblock CrimeForecaster: Crime Prediction by Exploiting the Geographical
  Neighborhoods’ Spatiotemporal Dependencies.
\newblock In \emph{The European Conference on Machine Learning and Principles
  and Practice of Knowledge Discovery in Databases}, 52--67.

\bibitem[{Verikas, Gelzinis, and Bacauskiene(2011)}]{verikas2011mining}
Verikas, A.; Gelzinis, A.; and Bacauskiene, M. 2011.
\newblock Mining data with random forests: A survey and results of new tests.
\newblock \emph{Pattern recognition}, 44(2): 330--349.

\bibitem[{Wu et~al.(2020)Wu, Pan, Long, Jiang, Chang, and
  Zhang}]{wu2020connecting}
Wu, Z.; Pan, S.; Long, G.; Jiang, J.; Chang, X.; and Zhang, C. 2020.
\newblock Connecting the dots: Multivariate time series forecasting with graph
  neural networks.
\newblock In \emph{ACM SIGKDD Conference on Knowledge Discovery and Data
  Mining}, 753--763.

\bibitem[{Wu et~al.(2019)Wu, Pan, Long, Jiang, and Zhang}]{wu2019graph}
Wu, Z.; Pan, S.; Long, G.; Jiang, J.; and Zhang, C. 2019.
\newblock Graph WaveNet for Deep Spatial-Temporal Graph Modeling.
\newblock In \emph{Twenty-Eighth International Joint Conference on Artificial
  Intelligence {IJCAI-19}}.

\bibitem[{Xu et~al.(2020)Xu, Lin, Xia, Guo, and Li}]{xu2020sume}
Xu, F.; Lin, Z.; Xia, T.; Guo, D.; and Li, Y. 2020.
\newblock Sume: Semantic-enhanced urban mobility network embedding for user
  demographic inference.
\newblock \emph{Proceedings of the ACM on Interactive, Mobile, Wearable and
  Ubiquitous Technologies}, 4(3): 1--25.

\bibitem[{Yu, Yin, and Zhu(2017)}]{yu2017spatio}
Yu, B.; Yin, H.; and Zhu, Z. 2017.
\newblock Spatio-temporal graph convolutional networks: A deep learning
  framework for traffic forecasting.
\newblock \emph{arXiv preprint arXiv:1709.04875}.

\bibitem[{Zhang et~al.(2020)Zhang, Chang, Meng, Xiang, and
  Pan}]{zhang2020spatio}
Zhang, Q.; Chang, J.; Meng, G.; Xiang, S.; and Pan, C. 2020.
\newblock Spatio-temporal graph structure learning for traffic forecasting.
\newblock In \emph{Proceedings of the AAAI Conference on Artificial
  Intelligence}, volume~34, 1177--1185.

\bibitem[{Zheng et~al.(2020)Zheng, Fan, Wang, and Qi}]{zheng2020gman}
Zheng, C.; Fan, X.; Wang, C.; and Qi, J. 2020.
\newblock Gman: A graph multi-attention network for traffic prediction.
\newblock In \emph{Proceedings of the AAAI Conference on Artificial
  Intelligence}, volume~34, 1234--1241.

\bibitem[{Zhu et~al.(2020)Zhu, Yan, Zhao, Heimann, Akoglu, and
  Koutra}]{zhu2020beyond}
Zhu, J.; Yan, Y.; Zhao, L.; Heimann, M.; Akoglu, L.; and Koutra, D. 2020.
\newblock Beyond Homophily in Graph Neural Networks: Current Limitations and
  Effective Designs.
\newblock \emph{Conference on Neural Information Processing Systems (NeurIPS)},
  33.

\end{thebibliography}

\end{document}